\pdfoutput=1
\documentclass{article}

\usepackage{arxiv}
\usepackage{natbib}
\usepackage{natbib}
\bibpunct[, ]{(}{)}{,}{a}{}{,}%

\usepackage{hyperref}
\hypersetup{
    colorlinks,
    linkcolor={red!50!black},
    citecolor={blue!50!black},
    urlcolor={blue!80!black}
}

\usepackage{times}
\usepackage{latexsym}
\usepackage{booktabs}
\usepackage{multirow}
\usepackage{amsmath}
\usepackage{float}
\usepackage{subcaption}
\usepackage{comment}
\usepackage{array}
\usepackage[table, svgnames, dvipsnames]{xcolor}
\usepackage[T1]{fontenc}
\usepackage[utf8]{inputenc}

\usepackage{microtype}

\usepackage{inconsolata}

\usepackage{graphicx}

\title{Benchmarking Sociolinguistic Diversity in Swahili NLP: A Taxonomy-Guided Approach}

\author{
  Kezia Oketch, John P. Lalor, Ahmed Abbasi \\
  Department of IT, Analytics, and Operations \\
  University of Notre Dame \\
  \texttt{\{koketch, john.lalor, aabbasi\}@nd.edu}
}

\begin{document}
\maketitle

\begin{abstract}
We introduce the first taxonomy-guided evaluation of Swahili NLP, addressing gaps in sociolinguistic diversity. Drawing on health-related psychometric tasks, we collect a dataset of 2,170 free-text responses from Kenyan speakers.
The data exhibits tribal influences, urban vernacular, code-mixing, and loanwords. 
We develop a structured taxonomy and use it as a lens for examining model prediction errors across pre-trained and instruction-tuned language models. 
Our findings advance culturally grounded evaluation frameworks and highlight the role of sociolinguistic variation in shaping model performance.
\end{abstract}

\section{Introduction}
Natural language processing (NLP) has advanced key sectors such as healthcare \citep{de2020sentiment}, education \citep{Obiria2018SwahiliTC}, and public policy. These gains are largely confined to high-resource languages such as English, Mandarin, and French; low-resource languages have been severely underrepresented in training data and model development \citep{mohamed2023bantulm, alabi-etal-2022-adapting,akera2022machine}. As a result, NLP systems often do not reflect the linguistic and cultural realities of the communities they are deployed in \citep{Magueresse2020LowresourceLA}, especially in multilingual, culturally diverse regions.

Swahili, spoken by over 200 million people across East and Central Africa \citep{merritt1985swahili,muna2017international}, exemplifies this disparity. Despite its widespread use as a \textit{lingua franca} connecting over 40 ethnolinguistic communities in Kenya alone \citep{petzell2012linguistic,murindanyi2023explainable,lodhi1993language,amidu1995kiswahili}, Swahili remains marginalized in NLP research and benchmark datasets \citep{martin2021sentiment,gelas2012developments,Obiria2018SwahiliTC}. Existing
datasets, such as MasakhaNER \citep{adelani2021masakhaner}, have made
strides in curating resources for African languages. However, these datasets are
mostly designed for structured tasks using formal language corpora,
leaving gaps in their ability to capture the dynamic linguistic features present in real-world usage.

Swahili's real-world usage is shaped by rich sociolinguistic variation, which poses challenges for NLP models trained on standardized corpora. Speakers routinely switch between dialects, borrow from tribal lexicons, blend English with Swahili in urban vernaculars like \textit{Sheng}, and use loanwords from Arabic and other regional languages \citep{mazrui1998neo}. These variations are not random; they reflect underlying social dimensions such as tribal identity, education, and socio-economic status, all of which influence lexical choices and discourse styles. For NLP models, this variation complicates semantic disambiguation, contextual inference, and syntactic parsing \citep{van2022intersection,pereraexploring,kumar2024multilingual}. For instance, the word \textit{Panda} in Swahili can mean ``climb,'' ``plant,'' or ``board'' depending on context; the \textit{Sheng} expression \textit{Dawa za kutesa} (``serious medication'') is used to describe something ``tough'' or ``powerful.'' 
%Such forms are difficult to parse using token-based models alone. 
We address these gaps by introducing a new Swahili NLP dataset designed to capture
this full spectrum of variation. 
Focused on health-related perceptions, it comprises 2,170 free-text responses from Kenyan speakers, each paired with demographic metadata such as tribe, gender, income, and education, enabling sociolinguistic modeling and fairness evaluation.
The dataset includes examples of
\textit{Sheng}, dialectal Swahili, code-mixing, and loan words, as well as instances of polysemy and informal usage. 

% While prior work has examined dialectal variation and demographic fairness in NLP \citep{blodgett-etal-2020-language}, our study is the first, to our knowledge, to develop a structured taxonomy of Swahili sociolinguistic phenomena and integrate it into prediction error modeling. We contribute a culturally grounded dataset of health-related text responses enriched with demographic annotations; benchmark performance and fairness across multilingual PLMs and LLMs; and reveal how linguistic variation such as code-mixing, loanwords, and tribal lexicons, systematically shapes model behavior. By centering underrepresented linguistic forms, our work highlights the need for sociolinguistically-aware evaluation frameworks and offers a pathway toward more equitable language technologies.

Our contributions are as follows. 
First, we collect a dataset of health psychometrics in Swahili, the first to the best of our knowledge, by following the procedure of \cite{abbasi-etal-2021-constructing}.
Second, we benchmark target-language and multi-lingual NLP models on the dataset to determine how they perform.
Third, we conduct a detailed analysis of the data and develop a taxonomy of Swahili sociolinguistic features.
Fourth, we use regression analysis to model prediction error as a function of taxonomy-guided features to analyze how sociolinguistic variation contributes to model performance.\footnote{Our code and data is available at \url{https://github.com/nd-hal/swahili_psych_taxonomy}.}

\section{Related Work}

\subsection{African Languages in NLP}
The linguistic diversity of Africa presents distinct challenges for NLP.
Over 2,000 languages are spoken on the continent, many of which lack digitized resources and annotated corpora \citep{joshi-etal-2020-state}. Swahili is the most widely spoken African language and as such has received more attention in the NLP community; however, existing models still struggle with its dialectal variation and tribal lexicon \citep{alabi-etal-2022-adapting, adelani2021masakhaner}. While initiatives like AfriBERTa and Masakhane have advanced representation for African languages \citep{nekoto-etal-2020-participatory, adelani2021masakhaner}, they remain limited in modeling sociolinguistic variation such as code-mixing, urban youth vernacular, and tribal influence. In Swahili specifically, tribal affiliation has been shown to shape vocabulary and discourse patterns \citep{mazrui1998neo}. We extend this work by curating a dataset that captures context-rich, non-standard variation and benchmarking NLP/LLM model performance.

\subsection{Sociolinguistic Variations}

Linguistic variations such as code-mixing, dialectal shifts, and urban vernaculars are often treated as noise in mainstream NLP workflows, where preprocessing favors standardized forms and disregards heterogeneity \citep{blodgett-etal-2020-language, joshi-etal-2020-state,kantharuban-etal-2023-quantifying}. This approach marginalizes communities whose speech deviates from dominant norms. However, a growing body of work challenges this premise. Studies on code-switching \citep{khanuja-etal-2020-gluecos, piergallini2016word}, dialectal NLP \citep{blodgett-etal-2016-demographic, faisal2024dialectbench}, and low-resource benchmarks \citep{tan2019assessing}, argue for treating variation as a meaningful signal. \citep{ponti2019modeling} note that variation is governed by typological, cognitive, and social factors and should be explicitly modeled. Likewise, \citep{bafna-etal-2024-evaluating} treat linguistic perturbations not as noise but as systematic deviations along morphological, phonological, and lexical dimensions.
%These patterns are linked to both linguistic distance and social marginalization.
We build on this argument by creating a taxonomy of sociolinguistic features in Swahili and modeling their influence on prediction error.
%, a step not taken by prior work.

%\subsection{Intersectional Bias and Fairness Metrics}

%While fairness has been widely studied in NLP \citep{mehrabi2021survey}, it is often the case that datasets do not have enough demographic data to consider fairness and bias across multiple demographic characteristics.
%\citep{lalor2022benchmarking} provide a benchmarking framework that measures
%intersectional fairness using metrics such as Disparate Impact (DI) and Fairness
%Violation (FV) by evaluting exclusive categories (e.g., comparing young men to older women) and found bias increases as the number of demographics under consideration increases. 
%These metrics quantify
%the extent to which models allocate resources unequally across intersectional
%subgroups, such as older women or lower-income individuals. 

% \textbf{General comment: There are several papers where two different versions are citep (e.g., Joshi et al 2020a and b). There are also arxiv papers citep even though a published version exists. Clean this up, so for each paper only one version (the version of record) is citep.} \ko{this is resolved}

\section{Dataset Construction}

NLP has the potential to elucidate how language mediates healthcare access, comprehension, and trust, including among linguistically marginalized populations \citep{aracena2023development}. Accordingly, we develop a dataset of healthcare related text data in Swahili. We adapt the framework of \cite{abbasi-etal-2021-constructing} by translating and reproducing their annotation task. 
Specifically, we adapted established English-language survey items originally measuring four health-related attitudes: \textit{Anxiety Visiting a Doctor}, \textit{Subjective Health Literacy}, \textit{Health Numeracy}, and \textit{Trust in Doctor} \citep{abbasi-etal-2021-constructing, netemeyer2020health}.\footnote{Henceforth, we refer to these dimensions as \textit{Anxiety}, \textit{Literacy}, \textit{Numeracy}, and \textit{Trust}, respectively.} A certified translator produced Swahili versions, which were reviewed by two experts for fluency and cultural fit. Discrepancies were resolved by consensus.

%The framework was translated into
%Swahili for linguistic and cultural relevance, to enable its application within the Kenyan context. User-generated responses were obtained using the Swahili prompts detailed in Table~\ref{table:psychometric}.

\begin{table*}[!t]
\centering
\footnotesize
\begin{tabular}{lp{6.75cm}p{6.5cm}}
\toprule
\textbf{Task} & \textbf{English Prompt} &
\textbf{Swahili Translation} \\ \midrule
Literacy & Regarding all the questions you
just answered, to what degree do you feel you have the capacity to obtain,
process, and understand basic health information and services needed to make
appropriate health decisions? Please explain your answer in a few sentences. 
& Kuhusiana na maswali uliyojibu hapo juu, kwa
kiwango gani unahisi una uwezo wa kupata, kuchakata, na kuelewa taarifa za
msingi za afya na huduma zinazohitajika kufanya maamuzi sahihi ya afya?
Tafadhali eleza jibu lako kwa sentensi chache. \\
Trust & In
a few sentences, please explain the reasons why you trust or distrust your
primary care physician. If you do not have a primary care physician, please
answer in regard to doctors in general.  & Katika sentensi chache, tafadhali eleza sababu kwa
nini unamwamini au humwamini daktari wako wa matibabu ya kawaida. Ikiwa huna
daktari wa matibabu ya kawaida, tafadhali jibu kuhusu madaktari kwa ujumla.\\ 
Anxiety & In a few sentences, please describe what makes you most anxious or worried when
visiting the doctor’s office. & Katika sentensi chache, tafadhali eleza
kinachokufanya uhisi wasiwasi au hofu zaidi unapotembelea ofisi ya daktari.\\
Numeracy & In a few sentences, please describe an experience in your life that demonstrated your knowledge of health or medical issues. & Kwa sentensi chache, tafadhali eleza tukio
ambalo lilionyesha ujuzi wako wa masuala ya afya au matibabu. \\ \bottomrule
\end{tabular}
\caption{English prompts and their corresponding Swahili translations for the four psychometric survey tasks} 
\label{table:psychometric}
\end{table*}

% \subsection{Dataset Creation and Summary Statistics}

For data collection, we utilized GeoPoll, a leading crowdsourcing platform in Kenya.
GeoPoll's robust network of participants in the region allowed
us to access a diverse demographic pool, capturing variations in Swahili usage
influenced by tribe and urban-rural divides. 
GeoPoll’s
mobile-first approach was crucial for engaging participants in rural and urban
areas, ensuring inclusivity in a region where internet access can be limited.

% This approach ensured that the data collection process
% was both efficient and representative of Swahili language use across different
% sociolinguistic contexts. 

Participants completed open-ended textual prompts (Table \ref{table:psychometric}) in addition to survey items corresponding to the four dimensions \citep{abbasi-etal-2021-constructing}. These included both self-reported behavior (e.g., frequency of doctor visits, smoking and drinking habits) and standard demographic questions (gender, age, tribe, education, income, and race).
%(\jpl{what behavioral questionaires?}\ko {added example behavioral questions}). 
To ensure response quality, we applied standard reliability practices \citep[e.g., incentives, validation, review,][]{abbasi-etal-2021-constructing, buechel-etal-2018-modeling,buhrmester2011amazon}. Participants received monetary compensation upon completion, consistent with local-fair wage guidelines. All study procedures were reviewed and approved by the University's Institutional Review Board, and informed consent was obtained from all participants.

% \jpl{Participants were paid...}

% \jpl{IRB information...}\ko {Information added}

\begin{figure*}
\begin{subfigure}[b]{0.45\textwidth}
    \centering
    \includegraphics[width=\linewidth]{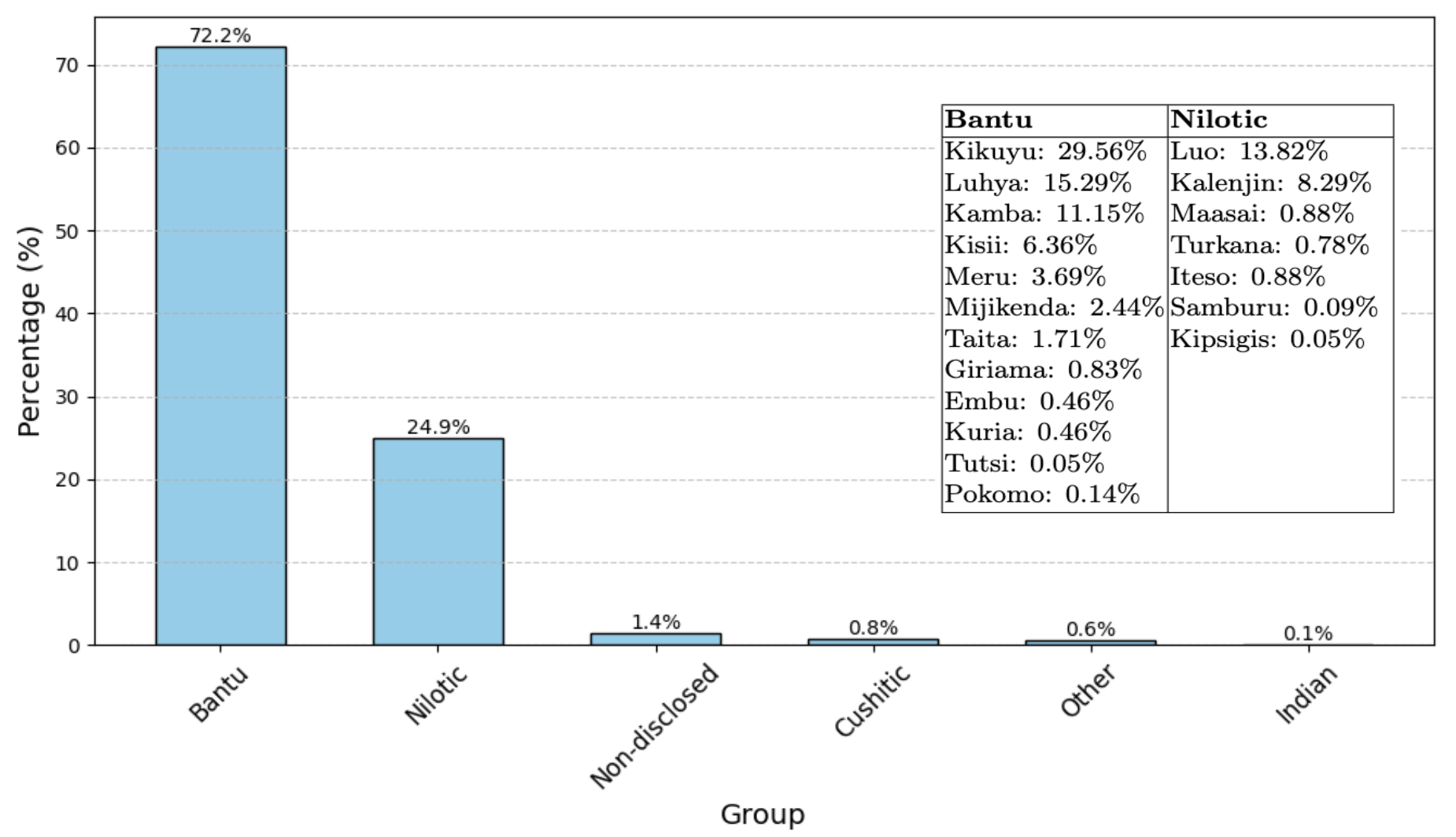}
    \caption{Distribution of tribes by language group.}
    \label{fig:tribe_distribution}
\end{subfigure}
~
\begin{subfigure}[b]{0.45\textwidth}
    \centering
    \includegraphics[width=\linewidth]{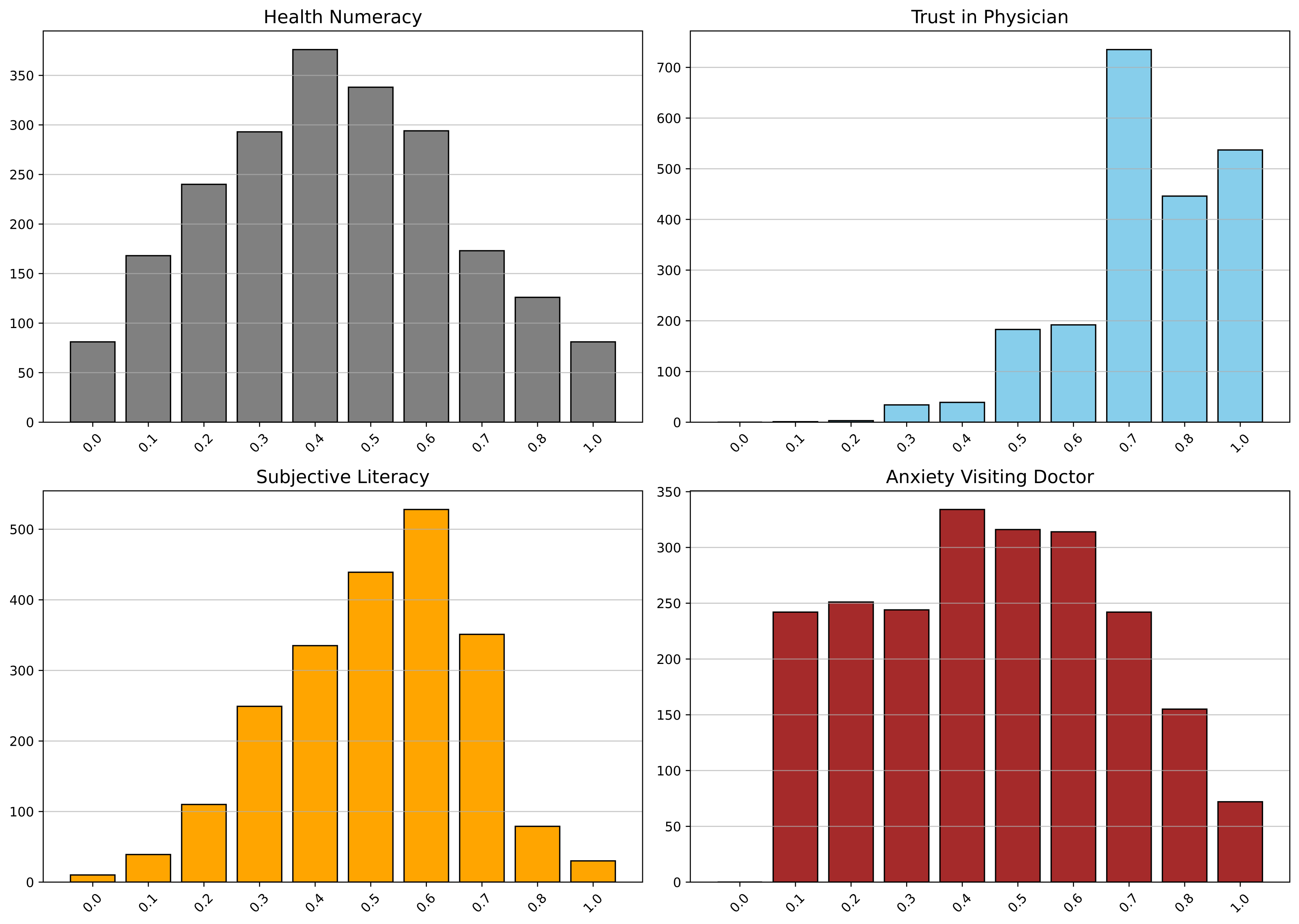}
    \caption{Distribution of psychometric task response scores.}
    \label{fig:score_distribution}
\end{subfigure}
\caption{Summary statistics of our collected dataset.}
\end{figure*}

To refine our survey design and validate the GeoPoll platform
we conducted a pilot study of
125 participants. 
%This pilot enabled us
%to refine our survey design and validate that GeoPoll could effectively capture the linguistic diversity we aimed to study. 
Following the pilot, we conducted our full study.
The final corpus comprises 2,170 free-text entries from native
speakers across Kenya.\footnote{Full summary statistics are provided in Appendix~\ref{sec:summStats}.} 
Since race is homogeneous ($>$95\% Black) in our testbed, tribe serves as a more salient demographic dimension, correlating with lexical choices.
Figure~\ref{fig:tribe_distribution} depicts the distribution of tribes in our
dataset grouped by their ethnolinguistic category. To quantify each construct, we aggregated item-level responses into a unified score scaled on a 0-1 range across tasks. Figure~\ref{fig:score_distribution} visualizes the distribution of user patterns across the four dimensions. We can see that numeracy and anxiety responses exhibit characteristics closer to normality, while trust and literacy deviate significantly due to skewness. Table~\ref{table:combined_av_scores} presents sample anxiety-related responses in Swahili, accompanied by their English translations and associated scores. The text was translated directly to English by a member of the research team to maintain the original respondent's intention. %\jpl{by a member of the research team(?)} \ko{Yes}

\begin{table*}[!htbp]
\centering
\small
\begin{tabular}{p{1.25cm}p{6.5cm}p{6.5cm}}
\toprule
\textbf{Anxiety\newline Score} & \textbf{Swahili Text Response} & \textbf{English
Translation} \\ \midrule
0.9524 & Pesa kwa matibabu hunifanya niwe na wasiwasi. Ripoti ya daktari hunipa
wasiwasi mwingi maanake sijui atasema naugua ugonjwa upi. & Money for treatment
makes me anxious. A doctor's report gives me a lot of worry because I don't know
which illness they will say I have. \\ 

0.3095 & Wasiwasi hutokea wakati usipojua shida au tatizo la afya nililonalo au
kutojua gharama ya matibabu pamoja na dawa kwa ujumla baada ya vipimo. & Anxiety
arises when you don't know the health problem or condition I have or not knowing
the overall cost of treatment and medication after tests. \\ \bottomrule
\end{tabular}
\caption{Sample participant responses to the Anxiety task, presented with corresponding Swahili text, English translation, and normalized psychometric scores.}
\label{table:combined_av_scores}
\end{table*}

\section{Benchmarking Swahili NLP Models}

\subsection{Experimental Setup}

To evaluate our Swahili dataset, we conducted regression and
classification tasks to see how well various multilingual NLP models could
predict survey-based ``ground-truth'' scores using the user-generated text responses \citep{abbasi-etal-2021-constructing}. 
In the regression setting, the target variable consisted of continuous scores normalized on a 0-1 range. For classification, we discretized across the median values, resulting in high versus low class labels \citep{gibson2015predicting,abbasi-etal-2021-constructing}. 
%For both tasks, we employed five-fold cross-validation. 

% \jpl{Add citations for each model in the next sentence.} \ko{Citations added}
We ran our experiments with nine NLP models: linear/logistic regression, mBERT \citep{devlin-etal-2019-bert}, XLM-RoBERTa \citep{conneau-etal-2020-unsupervised}, AfriBERTa \citep{ogueji-etal-2021-small}, SwahBERT \citep{martin-etal-2022-swahbert}, Qwen2.5-72B-Instruct and Qwen2.5-7B-Instruct \citep{yang2024qwen2}, and Llama-3.1-405B-Instruct and Llama-3.1-8B-Instruct \citep{grattafiori2024llama}.
The linear and logistic baselines were trained using TF-IDF features with 1-3-gram windows, and capped at 30,000 and 10,000 features, respectively. For PLMs, we fine-tuned each model for each task with five-fold cross validation using weighted cross-entropy loss for classification and mean squared error loss for regression. 
For instruction-tuned LLMs, we used few-shot prompting with three in-context examples sampled from the training fold. Prompts were structured to elicit either a numeric score between 0 and 1 for regression tasks, or a binary ``Yes'' or ``No'' for classification tasks. Additional experimental details, including hyperparameters and prompting templates, are in Appendix~\ref{sec:appendix B}.

% \jpl{more details here...} \ko{Details added}
%The details of the experiments are described in Appendix~\ref{sec:appendix B}

\subsection{Evaluation Metrics}

%\jpl{A few sentences on predictive performance metrics here...}
%We evaluated predictive performance using metrics appropriate to each task type. 
For regression tasks, we computed mean squared error (MSE) to quantify average prediction error and Pearson’s $r$ to capture the linear correlation between predicted and actual scores. For classification, we used the F1 score to balance precision and recall, and area under the ROC curve (AUC) to measure the model's ability to distinguish between classes across different thresholds.

% \jpl{add one sentence on why these four...} \ko{a sentence is added}
To evaluate tribal fairness, we selected a mix of multilingual (mBERT, XLM-RoBERTa), regional (AfriBERTa), and Swahili-specific (SwahBERT) models commonly used in African NLP tasks.\footnote{While prior research have evaluated NLP models with regards to race
\citep{guo2022auto}, in our dataset, race is homogeneous ($>$95\%
Black) and hence not a salient demographic variable for such analysis.} 
%We selected four pretrained language models (PLMs) - mBERT,
%XLM-RoBERTa, AfriBERTa, and SwahBERT - to evaluate fairness with regards to tribe.
%Tribe
%serves as a more meaningful demographic dimension, potentially correlating with lexical choices.
%These PLMs were chosen to represent 
We binarized the tribe variable such that the top two tribes (Kikuyu and Kamba) were the privileged class; all others were the
protected class. 

We quantify fairness via three measures: xAUC, Disparate Impact (DI), and Fairness Violation (FV). xAUC \citep{kallus2019fairness} measures disparities in the ranked ordering of predicted scores across groups: $\Delta \text{xAUC} = p(R_1^b \leq R_0^a) - p(R_1^a \leq R_0^b)$. A positive \( \Delta \mathrm{xAUC} \) indicates that individuals from group $a$ in the privileged class $T = 1$ are ranked higher by the model than those from group $b$ in the protected class $T = 0$. DI quantifies the relative rate at which two groups, privileged and non-privileged, receive predicted positive outcomes, serving as a proxy for group-level equity \citep{lalor-etal-2022-benchmarking,friedler2019comparative,barocas2016big} $DI = \frac{p(\hat{y} = 1 \mid T = 0)}{p(\hat{y} = 1 \mid T = 1)}$. 
%Given the true labels
%${y}$, predicted labels $\hat{y}$, and binary demographic attribute $T$, DI is defined as:
%.
%Here $T = 0$ denotes the protected group, while $T = 1$ represents the privileged group. 
A DI value below 1 indicates that the protected group receives positive predictions at a lower rate than the privileged group, signaling potential group-level disparity.
FV measures the largest absolute difference between the TPR of any intersectional subgroup and the global TPR, and quantifies the maximum discrepancy between the behavior of any subgroup and that of the overall population~\citep{yang2020fairness}. 
%When instantiated with the true positive rate (TPR), . % \jpl{We will get a question about why this binarization is different than the previous one. So we need a sentence here stating why we did it this way. Sticking with Kikyu, Luhya, and Kamba as our protected class would make sense here, as it's approx. 55\% of our data as opposed to over 70\%...} \ko{I added a note about the binarization but I will try the analysis with the top 3 tribes if we have time}

We also evaluated intersectional fairness across combinations of demographic attributes. Here, we binarized tribe into two ethnolinguistic groups, Bantu and Non-Bantu, to reflect established East African linguistic family structures and capture meaningful sociolinguistic variation \citep{mazrui1998power, heine2000african}.
%\footnote{We opted for this binarization rather than focusing on the top-2 tribes to ensure sufficient sample sizes within each intersectional subgroup, thereby preserving statistical power to detect disparities. While the Bantu group constitutes over 70\% of the dataset, we chose this binarization to balance linguistic validity with statistical power. Though this may introduce class imbalance, it avoids the extreme sparsity that would result from using granular tribal categories.}  
We computed DI and FV for all possible combinations of demographic attributes across increasing levels of complexity (e.g., two-way, three-way, and higher-order intersections). 
In the two-dimensional case, for example, protected groups include older women, women with lower income, and non-Bantu women; corresponding privileged groups are their complements, e.g., younger men, higher-income men, and Bantu men~\citep{lalor-etal-2022-benchmarking}. 
As in prior work, we set gender as the base demographic~\citep{bolukbasi2016man,
lalor-etal-2022-benchmarking}. 
We excluded 102 responses that contained ambiguous demographics  (e.g., cases where income was marked as ``not sure'' or tribe listed as ``I choose not to answer'').

\subsection{Results}

\subsubsection{Performance on Healthcare NLP Tasks}

\begin{table*}[ht]
\centering
\scriptsize
\setlength{\tabcolsep}{2pt}
\resizebox{\textwidth}{!}{%
\begin{tabular}{lcccccccccccccccc}
\toprule
\multirow{2}{*}{\textbf{Model}} 
& \multicolumn{8}{c}{\textbf{Continuous Classification}} 
& \multicolumn{8}{c}{\textbf{Binary Classification}} \\
\cmidrule(lr){2-9} \cmidrule(lr){10-17}
& \multicolumn{2}{c}{\textbf{Literacy}} & \multicolumn{2}{c}{\textbf{Trust}} & \multicolumn{2}{c}{\textbf{Anxiety}} & \multicolumn{2}{c}{\textbf{Numeracy}} 
& \multicolumn{2}{c}{\textbf{Literacy}} & \multicolumn{2}{c}{\textbf{Trust}} & \multicolumn{2}{c}{\textbf{Anxiety}} & \multicolumn{2}{c}{\textbf{Numeracy}} \\
& $r$ & RMSE & $r$ & RMSE & $r$ & RMSE & $r$ & RMSE
& AUC & $F1$ & AUC & $F1$ & AUC & $F1$ & AUC & $F1$ \\
\midrule
Regression & .098 & .388 & .288 & .157 & .123 & .261 & .022 & .297 & .603 & .589 & \textbf{.711} & \textbf{.784} & .558 & .570 & .563 & \textbf{.686} \\
AfriBERTa & .202 & .174 & .436 & \textbf{.133} & .156 & .234 & .129 & .242 & \textbf{.604} & .655 & .650 & .635 & \textbf{.585} & .647 & .570 & .648 \\
SwahBERT & .181 & .180 & .392 & .136 & .089 & .249 & .103 & .234 & .578 & .657 & .636 & .652 & .567 & .648 & .536 & .640 \\
XLM-RoBERTa & .145 & .172 & .398 & \textbf{.133} & .103 & .229 & .129 & \textbf{.207} & .572 & .659 & .647 & .669 & .557 & .652 & .525 & .636 \\
mBERT & .180 & .174 & .427 & \textbf{.133} & .138 & \textbf{.228} & \textbf{.133} & .220 & .559 & .659 & .649 & .652 & .543 & .652 & \textbf{.585} & .644 \\
Qwen2.5-7B & .121 & .215 & .337 & .250 & .084 & .266 & .031 & .294 & .538 & .565 & .566 & .496 & .483 & .386 & .527 & .572 \\
Qwen2.5-72B & .169 & .209 & .471 & .225 & .134 & .247 & .023 & .257 & .546 & .629 & .639 & .735 & .503 & .626 & .522 & .532 \\
Llama3.1-8B & .125 & \textbf{.118} & .163 & .187 & .039 & .255 & .019 & .212 & .533 & .526 & .563 & .665 & .503 & .607 & .523 & .591 \\
Llama3.1-405B & \textbf{.255} & .165 & \textbf{.541} & .215 & \textbf{.204} & .288 & .039 & .284 & .539 & \textbf{.675} & .608 & .712 & .517 & \textbf{.670} & .502 & .653 \\
\bottomrule
\end{tabular}
}
\caption{Performance of all evaluated models on regression and binary classification tasks across the four psychometric dimensions. The best-performing scores per column are highlighted in \textbf{bold}.}
\label{tab:performance}
\end{table*}

Table~\ref{tab:performance} summarizes performance on both regression and classification tasks. Our findings reveal two key insights. First, there is no clear consensus model (or category of models) that outperforms others. On certain tasks, such as Trust, domain-adapted PLMs consistently outperformed general-purpose LLMs (e.g., in terms of AUC, RMSE). For instance, AfriBERTa, XLM-RoBERTa, and mBERT all achieved the lowest RMSE on Trust ($.133$). AfriBERTa had the highest AUC on Literacy ($.604$) and Anxiety ($.585$), while mBERT led on Numeracy ($.585$), underscoring the advantage of multilingual models trained on regionally relevant data. Conversely, the two Llama LLMs had the best F1 scores for Anxiety and Literacy, and the highest correlation scores on Literacy ($r$ = .255), Trust ($r$ = .541), and Anxiety ($r$ = .204). 
%These results suggest that instruction tuning may help large models better capture subtle cues in open-ended text. However, their performance on classification tasks was less consistent, with relatively low AUC scores on Trust and Numeracy. This discrepancy may reflect a mismatch between few-shot prompting strategies and the linguistic complexity of Swahili, where morphological variation and lexical diversity pose challenges to generalization.

Second, when comparing the results against those attained by \cite{abbasi-etal-2021-constructing} on the English testbed, all PLM and LLM models in this study yielded AUCs for binary classification that were at least 20\%-30\% lower than those attained by their English-language BERT model (Literacy$=.798$, Trust$=.845$, Anxiety$=.723$, Numeracy$=.776$). This point is further underscored by the fact that in their study, the Regression models performed as weak baselines whereas in our results they were at or near the top for all four binary classification tasks. Although the two testbeds are obviously different (English versus Swahili), given the commonalities in the construction process, the results suggest that the state-of-the-art in regards to text classification predictive power for Swahili may be markedly behind that in high-resource languages such as English. 

%Third, traditional regression baselines, while outperformed by PLMs, exhibited surprising resilience in Trust (AUC=.711) and Numeracy (AUC=.686). These results suggest that n-gram-based models may still capture meaningful surface-level features, possibly due to lexical anchoring on high-frequency tokens.

%Overall, while scale enables some degree of zero-shot generalization, our findings reaffirm that linguistic and cultural adaptation remains essential for achieving robust performance in low-resource, sociolinguistically diverse settings. These trends are consistent with prior work on English-language psychometric data \citep{abbasi-etal-2021-constructing}, further validating the utility of culturally grounded benchmarks.

\subsubsection{LLMs and Domain-specific PLMs Encode Tribal Biases} %Fairness Results Reveal Tribal Disparities}
We first evaluated model fairness using the aforementioned xAUC metric, which is particularly appropriate for allocation-sensitive contexts such as triaging patients under limited healthcare resources, where ranking quality matters more than binary decisions \citep{kallus2019fairness}. For this analysis, in order to capture a mix of domain-specific PLMs and LLMs of different sizes, we included SwahBERT, Llama-3.1-405B, and Qwen2.5-72B.   
%xAUC \citep{kallus2019fairness} accounts for the ranked ordering of predicted scores. 
%This makes it particularly appropriate for allocation-sensitive contexts such as triaging patients under limited healthcare resources, where ranking quality matters more than binary decisions \citep{abbasi-etal-2021-constructing}. 
%xAUC between two demographic groups ($a$ and $b$) is defined as:
%\begin{equation}
%\begin{split}
%\Delta \text{xAUC} &= p(R_1^a > R_0^b) - p(R_1^b > R_0^a) \\
%&= p(R_1^b \leq R_0^a) - p(R_1^a \leq R_0^b)
%\end{split}
%\tag{1}
%\end{equation}
%A positive \( \Delta \mathrm{xAUC} \) indicates that individuals from group $a$ in the positive class $T = 1$ are ranked higher by the model than those from group $b$ in the negative class $T = 0$.

\begin{figure}[H]
    \centering
    \includegraphics[width=0.9\textwidth]{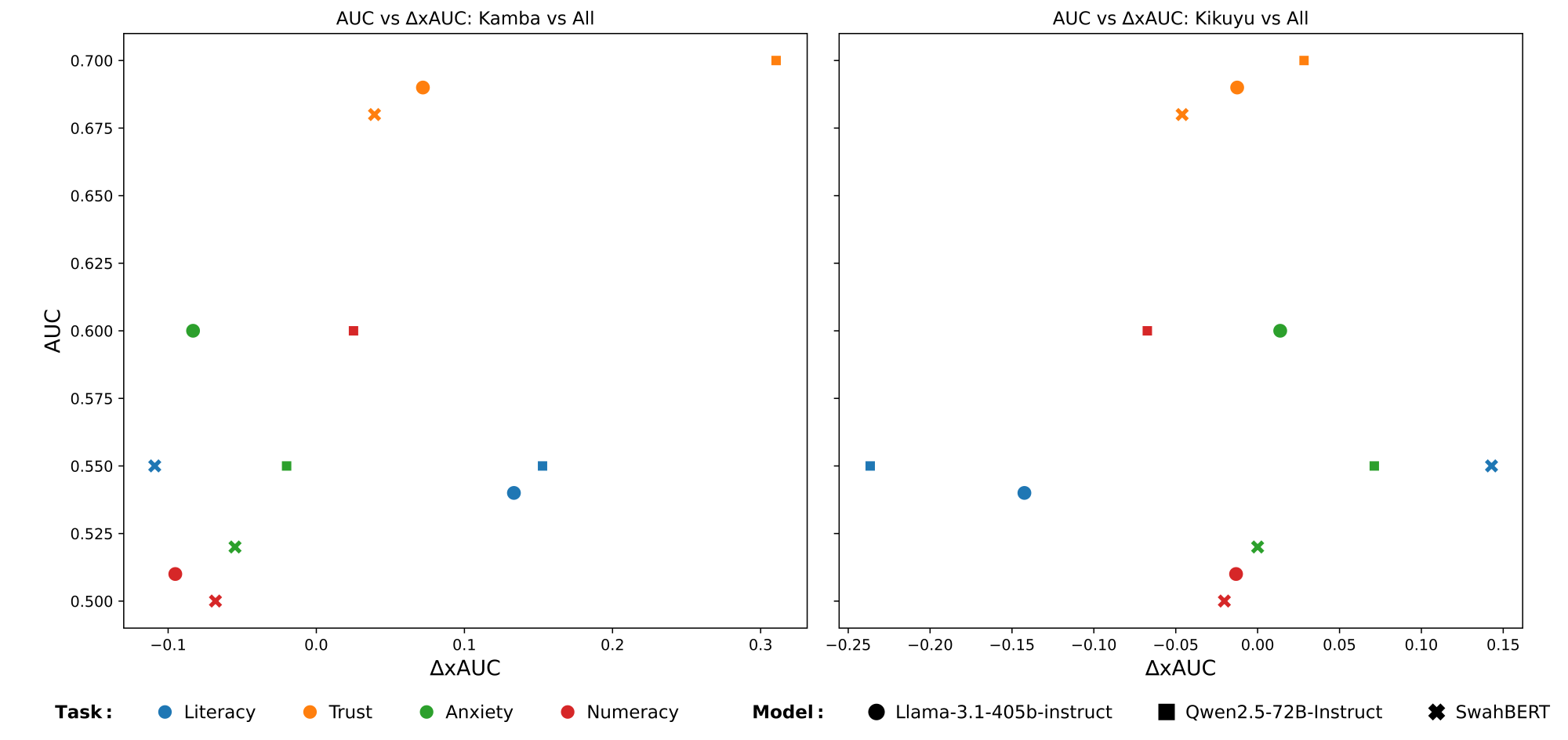} 
    \caption{Relationship between model performance (AUC) and fairness \( \Delta \mathrm{xAUC} \) for two tribal groups: (left) Kamba versus all others, and (right) Kikuyu versus all others.}
    \label{fig:kamba2-kiuk2}
\end{figure}

Figure~\ref{fig:kamba2-kiuk2} plots model performance (AUC) against $\Delta$xAUC, comparing Kamba vs All (left) and Kikuyu vs All (right). %While AUC captures model predictive , $\Delta$xAUC reflects whether members of a group are systematically ranked higher or lower than others, something critical in high-stakes domains like healthcare triage.  
The results reveal some strong tribal disparities. For instance, on the Kamba panel (left), the highest Trust AUC model, SwahBERT, exhibits the highest $\Delta$xAUC value ($\approx$ 0.3). This is a strong signal of ranking bias, indicating that the difference between Kamba individuals labeled as “trusting” over non-Kamba labeled as "distrusting," relative to non-Kamba "distrusting" over Kamba "trusting", is 30\%. Crucially, SwahBERT was also the only domain-specific PLM included in the analysis, and yet garnered markedley higher bias on the Trust classification task than its Llama and Qwen counterparts. 
%not the most accurate model, its AUC is around 0.60, while other models, Qwen2.5-72B and Llama-3.1-405B reach AUC scores closer to 0.70. This disconnect underscores a concerning dynamic: lower accuracy can coincide with greater bias. 
In this case, SwahBERT may be overfitting to lexical or syntactic features resulting in potential overgeneralizations about trust-related language used by Kamba speakers. We observe a similar trend on Literacy. This finding highlights that domain-adapted models can still entrench subgroup privilege when they mirror the linguistic norms of dominant communities. 

In the Kikuyu panel, we do not observe large $\Delta$xAUC deviations from 0 for the highest AUC models. However, the lower-performing models tend to exhibit large deltas on tasks such as Literacy indicating possible tribal disparities. 
%a negative skew for the majority of model–task combinations, indicating that Kikuyu individuals with positive labels are often ranked below non-Kikuyu individuals with negative labels, a pattern of systematic under-ranking. 
For instance, SwahBERT on Literacy reaches a $\Delta$xAUC of -0.24. While not all models exhibit this behavior, Qwen2.5-72B, for example, achieves a positive $\Delta$xAUC on Literacy ($\approx$ +0.14).
%, the overall trend suggests that Kikuyu individuals are either overlooked or deprioritized in ranking, despite valid positive predictions.
%Together, these disparities demonstrate that high performance does not imply equitable behavior. In both directions—over-ranking Kamba and under-ranking Kikuyu, 
The fairness results using xAUC suggest that both domain-specific PLMs and LLM models are encoding sociolinguistic patterns that systematically favor or disfavor certain tribal subgroups. In the ensuing subsection, we examine disparate impact across multiple demographics.

\subsubsection{Intersections Exacerbate Biases} 
\begin{figure}[H]
    \centering
    \includegraphics[width=0.9\textwidth]{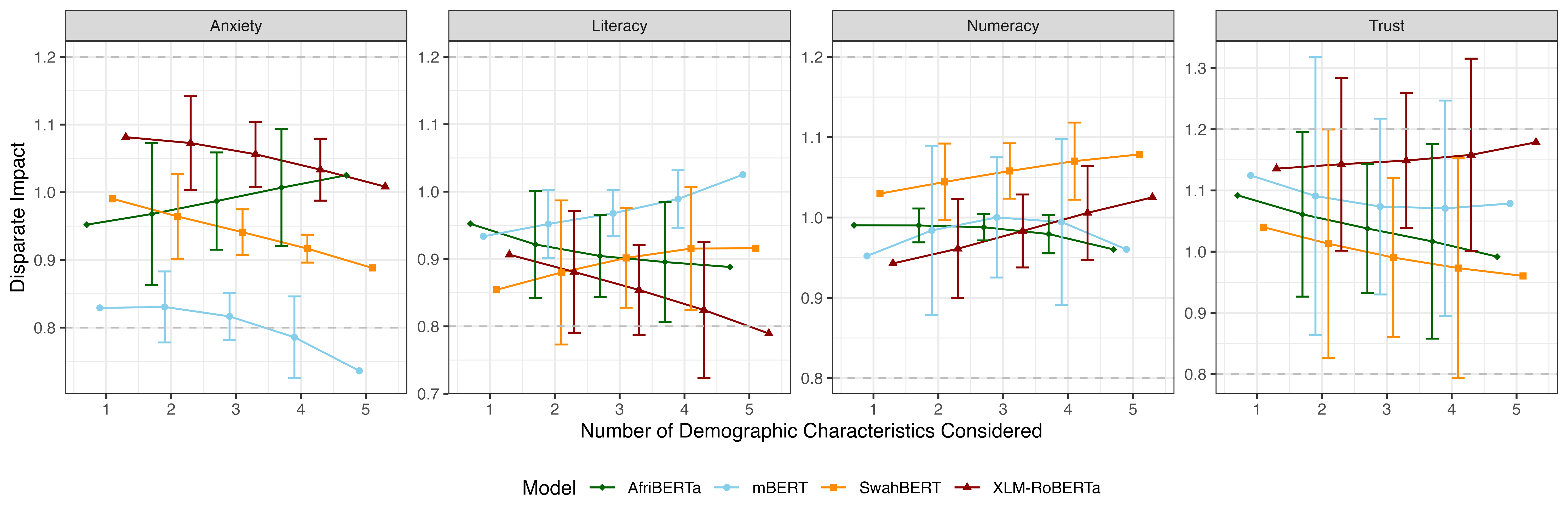}
    \caption{Impact of increasing demographic intersectionality on disparate impact (DI) across the four psychometric tasks. DI values are computed across 1-way to 5-way demographic intersections.}
    \label{fig:kf}
\end{figure}

Figure~\ref{fig:kf} shows results for disparate impact, moving from single-group fairness to intersectional subgroup analysis. The figure reveals that disparities intensify, and the apparent fairness of several models quickly erodes as the set of groups increases. On the Anxiety task, mBERT's disparate impact (DI) drops from  $\approx$0.83 at the group level to below 0.72 at five-way intersections, signaling that individuals belonging to multiple marginalized categories (e.g., old, non-Kamba, low-education, female, low-income) are systematically de-prioritized. XLM-RoBERTa and SwahBERT follow similar declines. 

In Literacy, XLM-RoBERTa's DI plummets below 0.80, revealing stark underprediction for complex identity configurations. These show that the more a person diverges from dominant demographic norms, the worse the model treats them. Numeracy appears more balanced (DI$\approx$1) for most model-intersection combinations.
%, but this obscures a deeper issue: even when performance is stable, representation may still be distorted, and subtle forms of overcompensation (e.g., DI>1 for SwahBERT) may reflect noise rather than fairness.
On the Trust task, the impact of the range of intersectional bias is most pronounced as evidenced by the growing intervals observed across all four PLMs. This trend is also observed on the Literacy task. Moreover, there, XLM-RoBERTa exhibits DI consistently<1 and below 0.8 when examining low-income, low-education, older females from select tribes. The results highlight key intersectional biases at the confluence of tribal affiliations and gender, age, education, and income.
%but that simply means that overprediction is now skewed in the other direction, introducing instability in subgroup treatment. These results make one thing clear: models don't just add up biases when someone belongs to multiple marginalized groups. They treat those individuals as if they're entirely unfamiliar, leading to unpredictable and often worse performance. A model that seems fair on one dimension can still fail badly when identities overlap. In culturally diverse, low-resource settings like ours, individuals routinely occupy multiple marginalized positions. If models cannot preserve fairness across these intersections, then they will fail the very users who are most at risk. Intersectional fairness auditing is not an optional deep-dive, it is the minimum threshold for responsible deployment.

% \jpl{Kezy, I think we can move Table 5 to an appendix, and focus on Figure 3. Please update Figure 3 to be a single row (with 4 columns) so it streches across the two columns.} \ko{updated figure 3 and moved table 5 to appendix A}

\section{Taxonomy of Linguistic Variation}

Having collected and evaluated our dataset with a benchmark suite of models, we next conduct a thorough analysis of the dataset to develop a taxonomy of linguistic variations in Swahili and conduct an error analysis based on the taxonomy.

\subsection{Taxonomy Construction}
\label{sec:taxonomy}
%Language models, particularly those pretrained on massive multilingual corpora, have significantly advanced NLP capabilities. 

As demonstrated by our benchmarking, PLMs and LLMs often struggle in culturally and linguistically rich contexts like Swahili, 
in particular when these models are deployed in region-specific applications \citep{grieve2024sociolinguistic,amol2024state}. 
These failures stem not only from data scarcity but also from the inability of PLMs to model sociolinguistic variations such as code-mixing, \textit{Sheng}, tribal lexical influence, and loanwords. 
These language patterns -- prevalent in everyday Swahili communication -- represent identity, region, and social contexts. 
%When models consistently fail to handle them correctly, it reflects not just technical limitations, but algorithmic bias and representational gaps \citep{blodgett-etal-2020-language, tan2019assessing}. 

%\jpl{How was the taxonomy constructed?}
To capture these linguistic patterns, we constructed a taxonomy through a bottom-up analysis of our data. We inspected all 2,170 responses from our survey across all four psychometric tasks. Each response was annotated by a member of the research team fluent in Swahili
%\jpl{by a member of the research team fluent in Swahili} 
for the presence of sociolinguistic features such as borrowed words, code-switched segments, and dialect-specific expressions. We grouped the most frequently observed patterns that commonly lead to errors in language modeling into four core categories: code-mixing, \textit{sheng}, tribal lexicon, and loanwords.\footnote{These four categories form the core of our analysis; they are not exhaustive. We also found other variations such as discourse markers, reduplication, and verbal morphology, that we do not model here (see Table \ref{tab:linguistic_features}). Here, we focus on the four categories most associated with model prediction errors.} 
% \jpl{Did we/how did we validate the coding? Did someone else review the annotation?} \ko{the annotations were not independently reviewed by a second coder}

\begin{figure*}[htbp]
    \centering
    \includegraphics[width=0.95\textwidth]{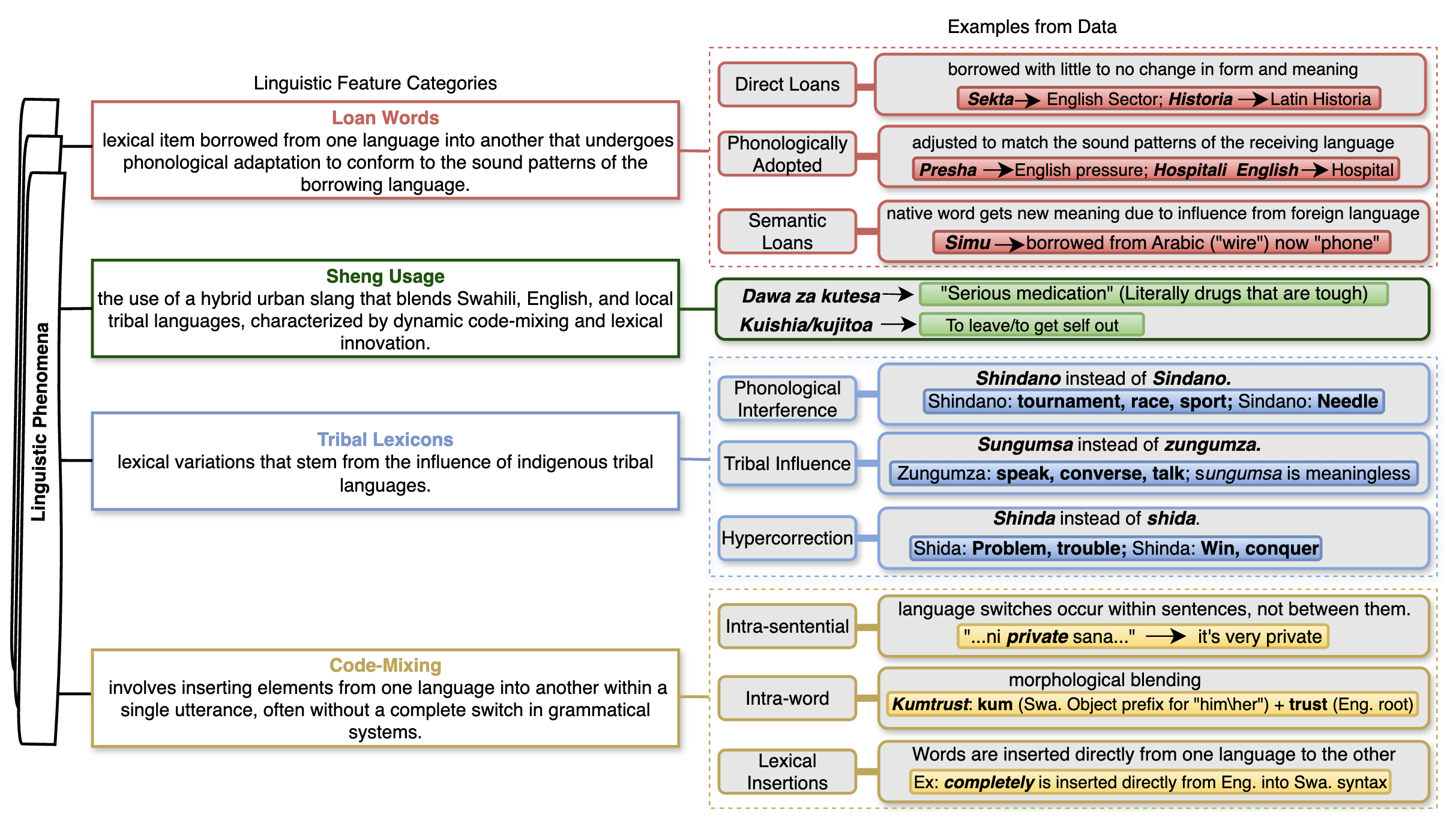}
    \caption{Proposed taxonomy of sociolinguistic variation in Swahili, capturing linguistic categories observed in the dataset: Loan Words \citep{kang2011loanword}, \textit{Sheng} \citep{githiora2002sheng}, Tribal Lexicons \citep{amol2024state}, Code-Mixing \citep{piergallini2016word}}
    \label{fig:taxonomy_diagram}
\end{figure*}

The final taxonomy (Figure~\ref{fig:taxonomy_diagram}) reveals specific representational gaps in current PLMs that have not been widely documented in prior literature. Through our error analysis of model predictions, we observed that the responses containing \textit{sheng} or tribal lexical terms were disproportionately affected. Some were consistently underpredicted, while others were overpredicted, depending on model and task. These inconsistencies suggest that PLMs lack robust representations for culturally embedded linguistic forms. By organizing these phenomena into a structured taxonomy, we offer a framework for identifying such weaknesses and improving model performance in linguistically diverse, underrepresented settings.

\subsection{Linguistic Features Like Sheng and Tribal Terms Drive Mispredictions}
Building on this taxonomy, we conducted a systematic error analysis to quantify how these linguistic features contribute to prediction errors made by the PLMs. Specifically, we shift focus from aggregate performance metrics to model behavior by analyzing prediction errors at the response level~\citep{lalor2024should}. Here we define error as the difference between the human-assigned and model-generated score (\(\Delta_i = \text{Human}_i - \text{Model}_i\)) and examine its variation by sociolinguistic feature. This enables empirical error analysis beyond anecdotal cases. As stated in~\ref{sec:taxonomy}, we annotated four sociolinguistic feature categories; code-mixing, \textit{sheng}, tribal lexicon, and loanwords. 
%(\jpl{Why these four??}).

We used the features to fit two ordinary least squares (OLS) regression models for each PLM-task pair. 
In the first model we consider main effects only: $\Delta_i = \beta_0 + \sum_{j=1}^{4} \beta_j X_{ij} + \varepsilon_i$, 
where \(X_{ij}\) is a binary indicator for whether linguistic feature \(j\) is present in response \(i\). 
This model estimates the average marginal effect of each feature on the model error. 
The second model includes all pairwise interactions among the four feature categories:

{
\begin{equation}
\Delta_i = \beta_0 + \sum_{j=1}^{4} \beta_j X_{ij} + \sum_{k>j} \beta_{jk}(X_{ij} \times X_{ik}) + \varepsilon_i
\label{eq:interaction_effects_compact}
\tag{3}
\end{equation}
}
%\noindent 
where \(X_{ij} \times X_{ik}\) represents the co-occurrence of features \(j\) and \(k\) in the same response. This captures interaction effects where feature combinations may influence model error. 
In both models, positive coefficients indicate underprediction by the model when a feature is present; negative values indicate overprediction. 

\begin{figure}[H]
    \centering
    \includegraphics[width=\textwidth]{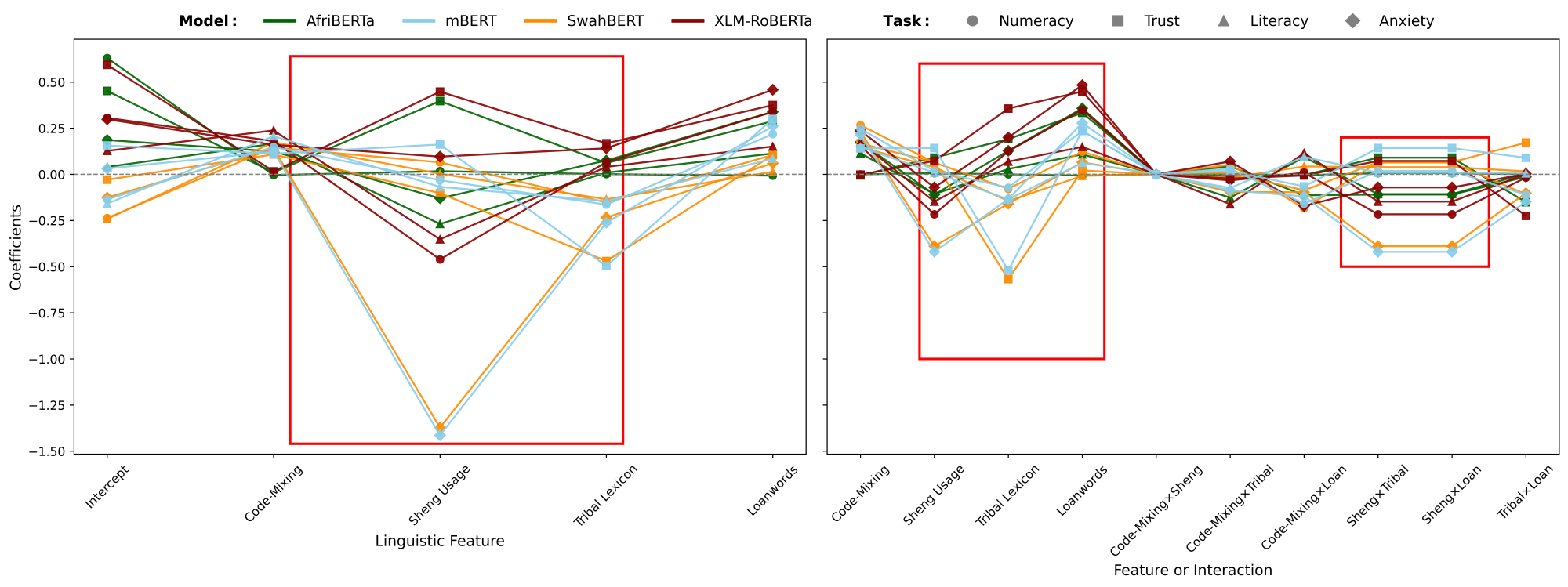}
    \caption{Ordinary least squares regression coefficients $\hat{\Delta}$ for main effects (left) and interaction effects (right) of sociolinguistic features on model prediction error across tasks.}
    \label{fig:m-e}
\end{figure}

The resulting coefficients \(\hat{\beta}_j\) are visualized in Figure~\ref{fig:m-e}. 
%The resulting coefficients are shown in the right side of Figure~\ref{fig:m-e}. 
Our findings show that model errors are not random but systematic. For instance, \textit{sheng} usage is consistently associated with errors; whether it is over-prediction or under-prediction is task-dependent. The interactions between \textit{sheng} usage, tribal lexicon, and loanwords are particularly strong indicators of under-prediction, especially in Numeracy and Anxiety (see area under red box). These results reflect broader trends in the literature: PLMs tend to struggle with code-mixed and dialectically marked language \citep{khanuja-etal-2020-gluecos, blodgett-etal-2016-demographic}, and such linguistic complexity often corresponds with higher error variance \citep{tan2019assessing}.

\begin{table}[H]
\centering
\small
\setlength{\tabcolsep}{4pt}
\begin{tabular}{llcccc}
\toprule
\textbf{Model} & \textbf{Feature} & \textbf{Anxiety} & \textbf{Literacy} & \textbf{Numeracy} & \textbf{Trust} \\
\midrule
\midrule
AfriBERTa & Sheng      & -0.130 (0.422)   & -0.269 (0.360)   & 0.017 (0.024)     & 0.397 (0.383)   \\
          & Tribal     & 0.075*** (0.022) & 0.009 (0.019)    & 0.001 (0.001)     & 0.061** (0.020) \\
          & Loan       & 0.339*** (0.018) & 0.113*** (0.016) & -0.007*** (0.001) & 0.288*** (0.016) \\
          & CodeMixed  & 0.124** (0.040)  & 0.166*** (0.034) & -0.004 (0.002)    & 0.010 (0.037)   \\
\midrule
mBERT     & Sheng      & -1.414** (0.433) & -0.068 (0.432)   & -0.032 (0.526)    & 0.162 (0.478)   \\
          & Tribal     & -0.262*** (0.022)& -0.145*** (0.023)& -0.164*** (0.028) & -0.497*** (0.023) \\
          & Loan       & 0.261*** (0.019) & 0.085*** (0.020) & 0.218*** (0.024)  & 0.300*** (0.021) \\
          & CodeMixed  & 0.123** (0.042)  & 0.207*** (0.041) & 0.148** (0.050)   & 0.112* (0.046)  \\
\midrule
SwahBERT  & Sheng      & -1.372** (0.418) & 0.003 (0.425)    & 0.065 (0.495)     & -0.102 (0.465)  \\
          & Tribal     & -0.234*** (0.022)& -0.135*** (0.022)& -0.153*** (0.026) & -0.470*** (0.023) \\
          & Loan       & 0.060** (0.019)  & 0.014 (0.020)    & 0.105*** (0.023)  & 0.104*** (0.021) \\
          & CodeMixed  & 0.140*** (0.040) & 0.174*** (0.041) & 0.145** (0.047)   & 0.108* (0.044)  \\
\midrule
XLM-RoBERTa & Sheng    & 0.096 (0.540)    & -0.352 (0.488)   & -0.461 (0.636)    & 0.448 (0.509)   \\
           & Tribal    & 0.141*** (0.029) & 0.040 (0.026)    & 0.066 (0.034)     & 0.168*** (0.027) \\
           & Loan      & 0.458*** (0.023) & 0.151*** (0.022) & 0.338*** (0.028)  & 0.375*** (0.022) \\
           & CodeMixed & 0.160** (0.052)  & 0.238*** (0.046) & 0.180** (0.061)   & 0.016 (0.049)   \\
\bottomrule
\end{tabular}
\caption{Main effect regression coefficients of sociolinguistic features on model prediction error, with standard errors shown in parentheses. Asterisks denote levels of statistical significance (* \(p<0.05\), ** \(p<0.01\), *** \(p<0.001\)).}
\label{tab:regression_coefs}
\end{table}
Importantly, these patterns have fairness implications. When models systematically mispredict responses that reflect certain linguistic identities, they contribute to epistemic injustice \citep{helm2024diversity}. By modeling $\Delta$ scores and surfacing linguistic conditions under which models deviate from human judgment, our analysis provides an interpretable framework for diagnosing model failure in low-resource languages.

\section{Conclusion}
This paper introduces a culturally grounded benchmark for evaluating fairness and performance for Swahili NLP. Based on the framework of \cite{abbasi-etal-2021-constructing}, we curated a novel dataset enriched with sociolinguistic variation and demographic metadata, and we developed a structural taxonomy to guide error analysis. Our experiments reveal that linguistic features such as \textit{sheng}, loanwords, and tribal lexicons systematically contribute to prediction errors. Furthermore, intersectional fairness metrics demonstrate that disparities compound across demographic dimensions such as tribe, gender, and education. 

Together, these findings underscore that fairness in NLP cannot be decoupled from sociolinguistic context. Our work offers not only a diagnostic tool for evaluating models in low-resource settings, but also a call to center linguistic diversity in the design of equitable language technologies.
%\jpl{This next para can maybe go here? } \ko{yes}
In the context of African NLP, intersectional bias remains underexplored. While multilingual models like mBERT \citep{devlin-etal-2019-bert} and XML-RoBERTa \citep{conneau-etal-2020-unsupervised} claim to generalize across languages, these models fail to capture intersectional patterns in African languages, particularly when tasked with representing tribal or gendered subgroups
\citep{joshi-etal-2020-state,nekoto-etal-2020-participatory}, highlighting the need for culturally grounded datasets.

%\newpage
\section*{Limitations}
While this study makes significant strides in advancing NLP research for
low-resource languages, particularly Swahili, it is important to acknowledge the
limitations that constrain its broader applicability. First, the dataset,
although rich in linguistic diversity, is primarily focused on Kenyan Swahili
speakers. This focus may limit the generalizability of the findings to other
regions where Swahili usage exhibits notable variations, such as Tanzania,
Uganda, and the Democratic Republic of Congo. Another challenge lies in the
dataset's representation of language phenomena. Despite the inclusion of \textit{sheng},
dialectal variations, and code-mixing, certain linguistic features, such as rare
idiomatic expressions and phonetic variations, are not comprehensively captured. Moreover, the study underexplores
code-switching involving Swahili and other regional languages, such as Kikuyu or
Luo, which are common in everyday communication. These limitations highlight the
inherent challenges in developing NLP systems for low-resource languages like
Swahili. They underscore the need for future research to expand datasets, refine
fairness evaluation methods, and develop models optimized for deployment in
resource-constrained environments. Additionally, although the cross-validation splits were stratified by task label, they were not demographically balanced by protected attributes (e.g., gender, tribe), which may have introduced representational bias during training. We highlight this as an important direction for future work.

% Entries for the entire Anthology, followed by custom entries

\bibliography{anthology,custom}
\bibliographystyle{acl_natbib}

\appendix 
\clearpage

\section{Appendix}
\label{sec:appendix A}
\subsection{Dataset Summary Statistics}
\label{sec:summStats}

% \begin{table}[h]
% \centering
% \footnotesize
% \begin{tabular}{p{2cm}p{4cm}}
% \toprule
% \textbf{Characteristic} & \textbf{Description} \\ \midrule
% Unique users & 2170 \\ 
% \multirow{4}{*}{\shortstack{Text fields \\(per user)}} & Literacy\\
% & Numeracy\\
% &Trust \\
% &Anxiety \\ 
% Age & Mean: 30.3, Over 50: 2.5\% \\ 
% \multirow{3}{*}{Race} & 95.3\% black\\
% &3.46\% multi-/bi-racial\\
% & 1.24\% other \\ 
% Sex (Male) & 52.72\% \\ 
% Income (KES) & 75.94\% $<$ 2.8 M \\ 
% Education & 72.73\% college grad or higher \\ 
% \multirow{5}{*}{Tribe (top 5)} & 29.59\% Kikuyu\\
% &15.3\% Luhya\\
% &13.82\% Luo\\
% &11.15\% Kamba\\
% &8.29\% Kalenjin \\ 
% \multirow{2}{*}{Response times} & 42 minutes (mean)\\
% & 39 minutes (median) \\ 
% Mean response lengths (chars) & 128 (Literacy), 121 (Numeracy),\newline 128 (Trust), 127 (Anxiety) \\ \bottomrule
% \end{tabular}
% \caption{Testbed summary statistics}
% \label{tab:dataset_summary}
% \end{table}

\begin{table}[h!]
\centering
\renewcommand{\arraystretch}{1.5} % Increase cell height for readability
\begin{tabular}{|>{\raggedright\arraybackslash}p{4cm}|p{8cm}|}
\hline
\textbf{Characteristic} & \textbf{Description} \\ \hline
Unique users & 2170 \\ \hline
Text fields (per user) & 
Literacy, 
Numeracy, 
Trust, 
Anxiety \\ \hline
Age & 
Mean: 30.3, 
Over 50: 2.5\% \\ \hline
Race & 
95.3\% black, 
3.46\% multi-/bi-racial, 
1.24\% other \\ \hline
Sex (Male) & 52.72\% \\ \hline
Income (KES) & 75.94\% $<$ 2.8 M \\ \hline
Education & 72.73\% college grad or higher \\ \hline
Tribe (top 5) & 
29.59\% Kikuyu, 
15.3\% Luhya, 
13.82\% Luo, 
11.15\% Kamba, 
8.29\% Kalenjin \\ \hline
Examples of other behavior/psychometric dimensions &
Usage of prescription drugs, 
Presence of primary care physician, 
Frequency of doctor visits, 
Smoking and drinking frequency \\ \hline
Response times & 
42 minutes (mean), 
39 minutes (median) \\ \hline
Mean response lengths (chars) & 
128 (Literacy), 121 (Numeracy), 
128 (Trust), 127 (Anxiety) \\ \hline
\end{tabular}
\caption{Summary statistics of the Swahili dataset, including participant demographics, text length, and survey completion details.}
\label{tab:dataset_summary}
\end{table}

\section{Appendix: Other Evaluation and Fairness Results}
\label{sec:appendix B}

\subsection{Model Evaluation}
Table~\ref{tab:merged} provides an expanded version of Table~\ref{tab:performance} that includes standard deviations calculated over five fold cross-validations. These results offer additional visibility into the stability of each model.

\begin{table*}[ht]
\centering
\scriptsize
\setlength{\tabcolsep}{2pt}
\resizebox{\textwidth}{!}{%
\begin{tabular}{lcccccccccccccccc}
\toprule
\multirow{2}{*}{\textbf{Model}} 
& \multicolumn{8}{c}{\textbf{Continuous Classification}} 
& \multicolumn{8}{c}{\textbf{Binary Classification}} \\
\cmidrule(lr){2-9} \cmidrule(lr){10-17}
& \multicolumn{2}{c}{\textbf{Literacy}} & \multicolumn{2}{c}{\textbf{Trust}} & \multicolumn{2}{c}{\textbf{Anxiety}} & \multicolumn{2}{c}{\textbf{Numeracy}} 
& \multicolumn{2}{c}{\textbf{Literacy}} & \multicolumn{2}{c}{\textbf{Trust}} & \multicolumn{2}{c}{\textbf{Anxiety}} & \multicolumn{2}{c}{\textbf{Numeracy}} \\
& $r$ & RMSE & $r$ & RMSE & $r$ & RMSE & $r$ & RMSE
& AUC & $F1$ & AUC & $F1$ & AUC & $F1$ & AUC & $F1$ \\
\midrule
Regression & .098$\pm$.074 & .388$\pm$.298 & .288$\pm$.054 & .157$\pm$.008 & .123$\pm$.019 & .261$\pm$.006 & .022$\pm$.041 & .297$\pm$.011 & .603$\pm$.015 & .589$\pm$.025 & \textbf{.711}$\pm$.022 & \textbf{.784}$\pm$.015 & .558$\pm$.017 & .570$\pm$.019 & .563$\pm$.021 & \textbf{.686}$\pm$.022 \\
AfriBERTa & .202$\pm$.049 & .174$\pm$.003 & .436$\pm$.029 & \textbf{.133}$\pm$.003 & .156$\pm$.043 & .234$\pm$.008 & .129$\pm$.033 & .242$\pm$.016 & \textbf{.604}$\pm$.028 & .655$\pm$.019 & .650$\pm$.022 & .635$\pm$.025 & \textbf{.585}$\pm$.017 & .647$\pm$.029 & .570$\pm$.029 & .648$\pm$.017 \\
SwahBERT & .181$\pm$.023 & .180$\pm$.012 & .392$\pm$.026 & .136$\pm$.004 & .089$\pm$.024 & .249$\pm$.009 & .103$\pm$.022 & .234$\pm$.004 & .578$\pm$.025 & .657$\pm$.005 & .636$\pm$.035 & .652$\pm$.025 & .567$\pm$.011 & .648$\pm$.016 & .536$\pm$.030 & .640$\pm$.024 \\
XLM-RoBERTa & .145$\pm$.035 & .172$\pm$.009 & .398$\pm$.011 & \textbf{.133}$\pm$.003 & .103$\pm$.042 & .229$\pm$.002 & .129$\pm$.064 & \textbf{.207}$\pm$.007 & .572$\pm$.021 & .659$\pm$.018 & .647$\pm$.024 & .669$\pm$.021 & .557$\pm$.013 & .652$\pm$.013 & .525$\pm$.048 & .636$\pm$.028 \\
mBERT & .180$\pm$.034 & .174$\pm$.009 & .427$\pm$.015 & \textbf{.133}$\pm$.002 & .138$\pm$.026 & \textbf{.228}$\pm$.004 & \textbf{.133}$\pm$.021 & .220$\pm$.012 & .559$\pm$.028 & .659$\pm$.011 & .649$\pm$.019 & .652$\pm$.022 & .543$\pm$.029 & .652$\pm$.013 & \textbf{.585}$\pm$.029 & .644$\pm$.029 \\
Qwen2.5-7B & .121$\pm$.033 & .215$\pm$.013 & .337$\pm$.027 & .250$\pm$.027 & .084$\pm$.033 & .266$\pm$.009 & .031$\pm$.058 & .294$\pm$.023 & .538$\pm$.022 & .565$\pm$.015 & .566$\pm$.023 & .496$\pm$.114 & .483$\pm$.013 & .386$\pm$.079 & .527$\pm$.009 & .572$\pm$.105 \\
Qwen2.5-72B & .169$\pm$.039 & .209$\pm$.012 & .471$\pm$.017 & .225$\pm$.014 & .134$\pm$.028 & .247$\pm$.008 & .023$\pm$.037 & .257$\pm$.024 & .546$\pm$.021 & .629$\pm$.012 & .639$\pm$.036 & .735$\pm$.059 & .503$\pm$.016 & .626$\pm$.029 & .522$\pm$.013 & .532$\pm$.093 \\
Llama3.1-8B & .125$\pm$.052 & \textbf{.118}$\pm$.014 & .163$\pm$.003 & .187$\pm$.003 & .039$\pm$.009 & .255$\pm$.023 & .019$\pm$.054 & .212$\pm$.019 & .533$\pm$.025 & .526$\pm$.041 & .563$\pm$.036 & .665$\pm$.019 & .503$\pm$.038 & .607$\pm$.077 & .523$\pm$.023 & .591$\pm$.129 \\
Llama3.1-405B & \textbf{.255}$\pm$.063 & .165$\pm$.013& \textbf{.541}$\pm$.042 & .215$\pm$.012 & \textbf{.204}$\pm$.025 & .288$\pm$.030 & .039$\pm$.059 & .284$\pm$.036 & .539$\pm$.023 & \textbf{.675}$\pm$.013 & .608$\pm$.053 & .712$\pm$.118 & .517$\pm$.016 & \textbf{.670}$\pm$.018 & .502$\pm$.006 & .653$\pm$.020 \\
\bottomrule
\end{tabular}
}
\caption{Expanded model performance metrics including standard deviations across five-fold cross-validation for both regression and binary classification tasks.}
\label{tab:merged}
\end{table*}

\subsection{Model Fairness}
\label{sec:appendix A.2}
\begin{figure*}[htbp]
    \centering
    \includegraphics[width=\textwidth]{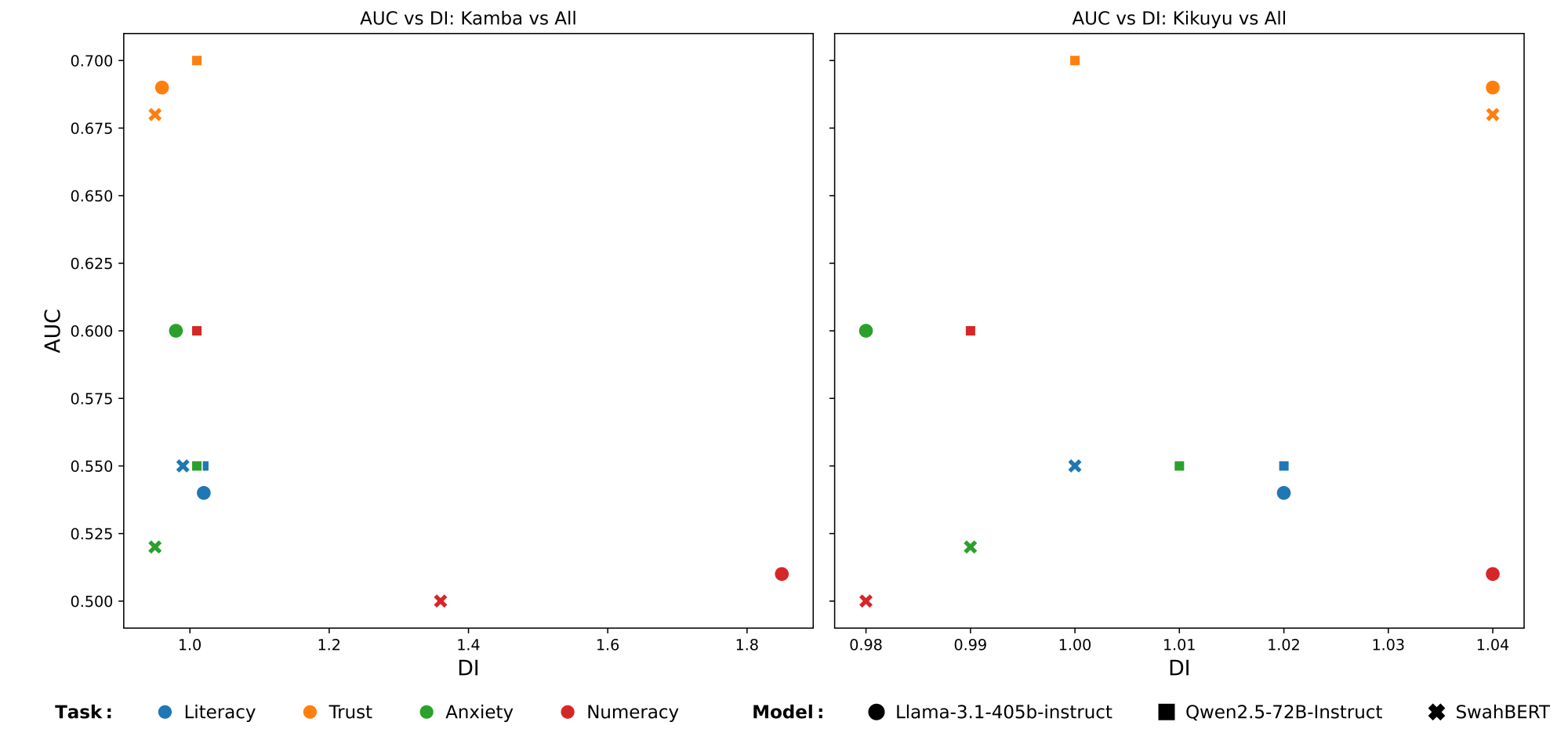} 
    \caption{Relationship between model performance (AUC) and group fairness, measured by Disparate Impact (DI), for the Kamba and Kikuyu groups compared against all other tribal subgroups.}
    \label{fig:kamba-kiuk}
\end{figure*}

We also assessed group-level fairness using DI, focusing on two tribes with high representation in our dataset: Kikuyu and Kamba. We computed DI for each task by comparing the positive prediction rates for these groups against the rest of the population (i.e., Kikuyu or Kamba as the privileged classes versus All other as the protected class). As shown in Figure~\ref{fig:kamba-kiuk}, the DI values for most tasks are close to 1, suggesting minimal disparity. However, in the Numeracy task for Kamba group, DI exceeds the 1.2 threshold when evaluated using Llama 3.1-405b and Qwen2.5-72B. 

This indicates that participants outside the Kamba group (i.e., the protected group) are more likely to receive positive predictions for Numeracy compared to Kamba participants. This disparity may stem from variation in how numeracy-related concepts are linguistically expressed across groups, with non-Kamba participants using phrasing that better aligns with the model’s learned patterns and trigger positive classifications.

\section{Appendix: Experiments and Model Parameters}
\label{sec:appendix B}

As mentioned in section 4, we conducted both regression and classification experiments across four psychometric tasks: Subjective Literacy, Trust in Physicians, Anxiety visiting the Doctor, and Health Numeracy. Each task involved predicting either a continuous score or a binarized label derived from human-annotated data. Our goal was to compare model performance and fairness across baseline models, PLMs, and LLMs in both zero-shot and few-shot settings.

\subsection{TF-IDF Baseline Models}
We implemented two classical baselines using scikit-learn: a linear regression model for continuous score prediction, and a logistic regression model for binary classification. For both, free-text responses were vectorized using TF-IDF with an n-gram range of 1 to 3. The logistic model used up to 30,000 features, while the linear model used up to 10,000 features to reduce dimensionality. Logistic regression was evaluated using Accuracy, F1, AUC, Precision, and Recall. Linear regression was evaluated using Root Mean Squared Error (RMSE) and Pearson correlation (r). All experiments used 5-fold cross-validation. We report the average F1, AUC, r, and RMSE across all five folds. 

\subsection{Pretrained Language Models}
We fine-tuned four transformer-based PLMs: mBERT, XLM-RoBERTa, AfriBERTa, and SwahBERT, using the Hugging Face Transformers library. For classification, we discretized continuous scores using the median and fine-tuned models with a batch size of 16, a learning rate of 1e-5, and 5 epochs. Weighted cross-entropy loss was used to mitigate class imbalance. For regression, we predicted the continuous scores directly using mean squared error loss, a batch size of 8, and otherwise similar hyperparameters. All PLMs were evaluated using 5-fold cross-validation on an NVIDIA A100 GPU. We report AUC and F1 scores for classification, and RMSE and Pearson r for regression.

\subsection{Large Language Models}
We evaluated four multilingual LLMs—Qwen2.5-72B-Instruct, Qwen2.5-7B-Instruct, Llama3.1-405B-Instruct, and Llama3.1-8B-Instruct—using few-shot prompting via the Replicate and DeepInfra APIs. For the regression task, each prompt included three few-shot examples (low, medium, high) from the training fold, followed by a test text, and instructed the model to return a numeric score between 0 and 1. For classification, prompts consisted of three binary-labeled examples and a test text, with the model instructed to respond with ``Yes'' or ``No.'' Predictions were parsed using basic string-matching logic.

All API calls used consistent decoding parameters across models: \texttt{temperature=0.3}, \texttt{max\_tokens=100}, \texttt{top\_p=1.0}, and no specified stop sequence. Each model received the same prompt format across folds. Evaluation was conducted per task using AUC and F1 for classification, and RMSE and Pearson’s $r$ for regression.

\begin{figure*}[ht]
    \centering
    \begin{minipage}[t]{0.48\linewidth}
        \centering
        \includegraphics[width=\linewidth]{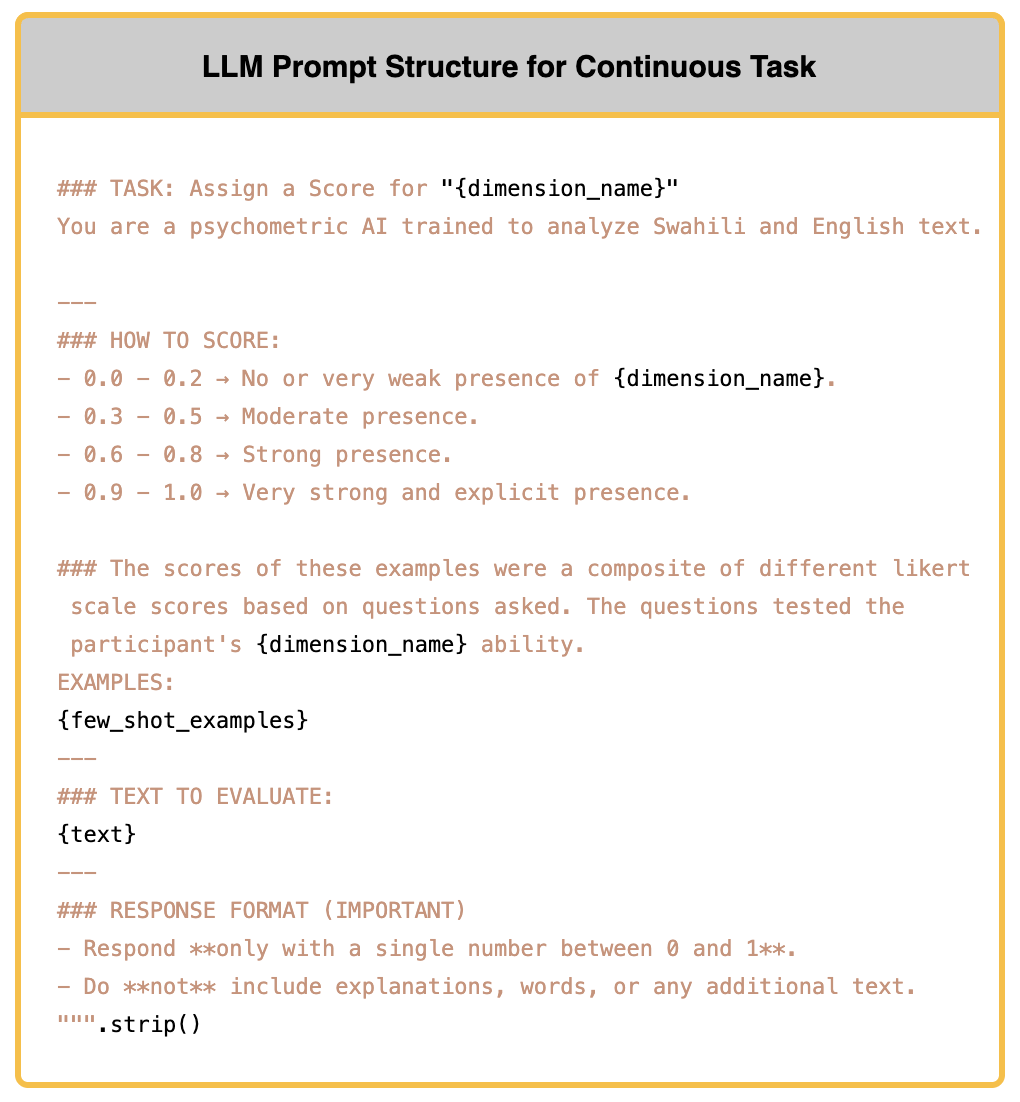}
        \caption*{(a) Few-shot Prompt} 
    \end{minipage}
    \hfill
    \begin{minipage}[t]{0.48\linewidth}
        \centering
        \includegraphics[width=\linewidth]{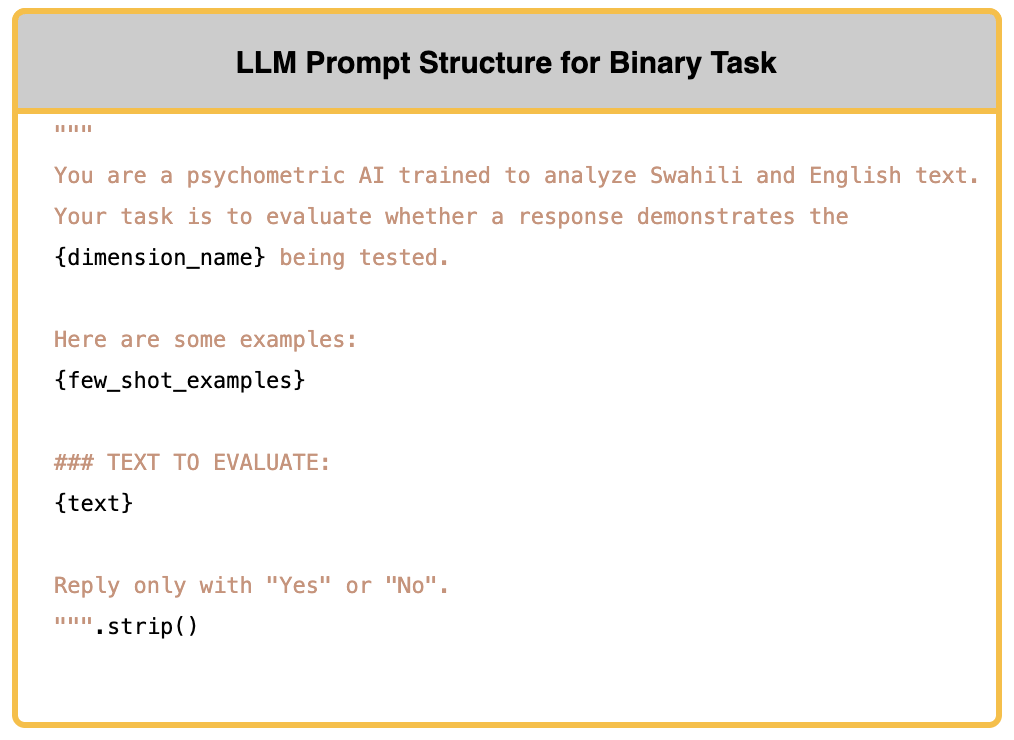}
        \caption*{(b) Binary Prompt}
    \end{minipage}
    \caption{Examples of prompt formats used in our experiments.}
    \label{fig:prompt_examples}
\end{figure*}

\subsection{Intersectional Fairness Evaluation}

We evaluated intersectional fairness by computing two complementary group-level metrics: \textit{Disparate Impact (DI)} and \textit{Fairness Violations (FV)}. Our analysis focused on five binary demographic attributes: age ($\leq$30), gender (male), tribal identity (Bantu vs. Non-Bantu), education ($\geq$ some college), and income ($\leq$ 2.58M KES). Each attribute was binarized to distinguish between a protected group (coded as 0) and a privileged group (coded as 1). To capture compounded disadvantage, we generated all 1-way to 3-way combinations of these attributes, resulting in 1, 2, and 3-way intersectional subgroups. Each combination was encoded as a binary indicator representing whether a participant belonged to all constituent protected categories. This allowed us to examine model performance disparities not only across individual demographic groups but also across their intersections. We excluded subgroups with fewer than two members in either protected or privileged categories to ensure statistical robustness. DI and FV metrics were aggregated across 1-way (DI), 2-way (DI\textsuperscript{+}, FV\textsuperscript{+}), and 3-way (DI\textsuperscript{++}, FV\textsuperscript{++}) levels and are summarized in Table~\ref{tab:int-table}.

While group-level fairness scores indicate relatively limited disparities, the intersectional analysis reveals that combining multiple demographic attributes surfaces more complex patterns of bias (Table ~\ref{tab:int-table}). A clear compounding effect emerges in the Anxiety task for mBERT, where DI decreases from 0.90 at the group level to 0.86 (DI+) and further to 0.83 (DI++). This pattern suggests that as more identity dimensions are layered, the model increasingly underpredicts for certain subgroups.  Fairness Violation (FV) scores further confirm this trend. In the Trust task, XLM-RoBERTa’s score increases from 0.60 at the group level (FV) to 0.80 at the three-way intersection level (FV++), reflecting a sharp rise in disparity as more demographic attributes are considered. These findings demonstrate that fairness deteriorates when multiple identity dimensions are considered jointly. 

%Intersectional Table
\begin{table}[!htbp]
\centering
\scriptsize
\setlength{\tabcolsep}{3pt}
\begin{tabular}{llrrrrrr}
\toprule
\textbf{Task} & \textbf{Model} & \textbf{DI} & \textbf{DI+} & \textbf{DI++} & \textbf{FV} & \textbf{FV+} & \textbf{FV++} \\
\midrule
\multirow{4}{*}{\textbf{Literacy}} 
& mBERT & 0.94 & 0.98 & 0.99 & 0.20 & - & - \\
& XLM-RoBERTa & \textbf{0.81} & \textbf{0.88} & \textbf{0.86} & 0.20 & \textbf{0.10} & \textbf{0.20} \\
& AfriBERTa & 0.93 & 0.94 & 0.93 & - & - & - \\
& SwahBERT & 0.93 & 0.95 & 0.94 & \textbf{0.40} & \textbf{0.10} & - \\
\midrule
\multirow{4}{*}{\textbf{Trust}} 
& mBERT & 1.05 & 1.04 & 1.05 & \textbf{0.60}  & 0.40 & 0.30 \\
& XLM-RoBERTa & \textbf{1.19} & \textbf{1.15} & \textbf{1.16} & \textbf{0.60} & \textbf{0.70} & \textbf{0.80} \\
& AfriBERTa & 1.05 & 1.02 & 1.01 & 0.20 & 0.20 & 0.20 \\
& SwahBERT & 1.03 & 1.00 & 0.99 & 0.40 & 0.20 & 0.30 \\
\midrule
\multirow{4}{*}{\textbf{Anxiety}} 
& mBERT & \textbf{0.90} & \textbf{0.86} & \textbf{0.83} & - & - & \textbf{0.30} \\
& XLM-RoBERTa & 1.09 & 1.07 & 1.06 & \textbf{0.20} & \textbf{0.20} & 0.10 \\
& AfriBERTa & 0.96 & 0.97 & 0.99 & - & - & - \\
& SwahBERT & 0.95 & 0.96 & 0.94 & - & - & - \\
\midrule
\multirow{4}{*}{\textbf{Numeracy}} 
& mBERT & 1.07 & 1.02 & 1.00 & \textbf{0.40} & \textbf{0.20} & \textbf{0.10} \\
& XLM-RoBERTa & \textbf{0.94} & 0.96 & 0.98 & - & - & - \\
& AfriBERTa & 0.96 & \textbf{0.95} & \textbf{0.95} & - & - & - \\
& SwahBERT & 1.03 & 1.02 & 1.02 & 0.20 & - & - \\
\bottomrule
\end{tabular}

\caption{Disparate Impact (DI) and Fairness Violation (FV) scores for one-way (DI, FV), two-way (DI$^+$, FV$^+$), and three-way (DI$^{++}$, FV$^{++}$) intersectional subgroup analyses. Bolded values indicate the greatest observed disparity. Dashes indicate zero values.} 
\label{tab:int-table}
\end{table}

\textbf{Note on Tribe Binarization:} Tribal identity was binarized as \textit{Bantu vs. Non-Bantu} to reflect East African linguistic family structures \citep{mazrui1998neo, heine2000african}. Bantu languages, such as Kikuyu and Kamba, share agglutinative morphology and phonological structures, while Non-Bantu groups, such as Luo and Somali, come from Nilotic and Cushitic families with distinct syntactic and lexical properties. This binarization captures meaningful sociolinguistic variation while preserving statistical power and interpretability in fairness evaluations.
\section{Appendix: Additional Dataset Information}
\label{sec:appendix C}
The table below summarizes the frequency of linguistic features identified in the Swahili dataset. The counts represent the total number of instances each feature type appeared across all text responses. This taxonomy guided our fairness analysis by enabling targeted evaluation of model performance across diverse linguistic phenomena.

\begin{table}[htbp]
    \centering
    \footnotesize
    \renewcommand{\arraystretch}{1.1}
    \setlength{\tabcolsep}{4pt}
    \begin{tabular}{p{2.8cm}|p{4.8cm}r}
        \toprule
        \textbf{Feature Group} & \textbf{Category} & \textbf{Count} \\
        \midrule
        Lexical & Vocabulary Richness (Hapax Legomena) & 145{,}628 \\
        \cline{2-3}
               & Word Length Variation & 360 \\
        \cline{2-3}
               & Loan Words & 2{,}819 \\
        \cline{2-3}
               & Code-Mixing & 245 \\
        \cline{2-3}
               & Sheng & 17 \\
        \cline{2-3}
               & Function Word Usage & 347{,}448 \\
        \midrule
        Structural & Word Count \& Distribution & 175{,}269 \\
        \cline{2-3}
                  & Reduplication & 245 \\
        \cline{2-3}
                  & Absence of Phonetic Markers & 11 \\
        \cline{2-3}
                  & Punctuation Frequency & 10{,}938 \\
        \cline{2-3}
                  & Question/Exclamation Usage & 31 \\
        \midrule
        Syntactic & Verb Morphology & 9{,}230 \\
        \cline{2-3}
                 & Discourse Markers & 2{,}755 \\
        \midrule
        Content-Specific & Medical Terminology & 8{,}274 \\
        \cline{2-3}
                        & Tribal Lexicon & 2{,}323 \\
        \bottomrule
    \end{tabular}
    \caption{Frequency of Linguistic Features in the Dataset}
    \label{tab:linguistic_features}
\end{table}

% \bibliographystyle{plainnat}
% \bibliography{custom}
\end{document}